# Bengali to Assamese Statistical Machine Translation using Moses (Corpus Based)

Nayan Jyoti Kalita[1], Baharul Islam[2]
[1]Department of CSE, Royal School of Engineering and Technology
[2]Department of IT, Gauhati University
Guwahati, India
{nayan.jk.123, islambaharul65}@gmail.com

*Abstract*—Machine dialect interpretation assumes a real part in encouraging man-machine correspondence and in addition men-men correspondence in Natural Language Processing (NLP). Machine Translation (MT) alludes to utilizing machine to change one dialect to an alternate. Statistical Machine Translation is a type of MT consisting of Language Model (LM), Translation Model (TM) and decoder. In this paper, Bengali to Assamese Statistical Machine Translation Model has been created by utilizing Moses. Other translation tools like IRSTLM for Language Model and GIZA-PP-V1.0.7 for Translation model are utilized within this framework which is accessible in Linux situations. The purpose of the LM is to encourage fluent output and the purpose of TM is to encourage similarity between input and output, the decoder increases the probability of translated text in target language. A parallel corpus of 17100 sentences in Bengali and Assamese has been utilized for preparing within this framework. Measurable MT procedures have not so far been generally investigated for Indian dialects. It might be intriguing to discover to what degree these models can help the immense continuous MT deliberations in the nation.

## I. INTRODUCTION

Multilingualism is considered to be a part of democracy. With increasing growth of technology language barrier should not be a problem. It becomes important to provide information to people as and when needed as well as their native language. Machine translation is not primarily an area of abstract intellectual inquiry but the application of computer and language sciences to the development of system answering practical needs. The focus of the research presented here was to investigate the effectiveness of a phrase based statistical Bengali-Assamese translation using the Moses toolkit.

The field of common dialect handling (NLP) started give or take five decades prior with machine interpretation frameworks. In 1946, Warren Weaver and Andrew Donald Booth examined the specialized attainability of machine interpretation "by method for the methods created throughout World War II for the breaking of adversary codes" [1]. Throughout the more than fifty years of its presence, the field has developed from the lexicon based machine interpretation frameworks of the fifties to the more versatile, powerful, and easy to use NLP situations of the nineties.

**Machine Interpretation**

Machine interpretation is the name for modernized systems that mechanize all or some piece of the procedure of making an interpretation of starting with one dialect then onto the next. In a huge multilingual public opinion like India, there is incredible interest for interpretation of records starting with one language then onto the next language. There are 22 intrinsically sanction languages, which are authoritatively utilized as a part of distinctive states. There are something like 1650 tongues talked by distinctive groups. There are 10 Indict scripts. These dialects are overall created and rich in substance. They have comparative scripts and sentence structures. The alphabetic request is likewise comparable. A few dialects use regular script, particularly Devanagari. Hindi composed in the Devanagari script is the official language of the Government of India. English is likewise utilized for government notices and interchanges. India's normal writing proficiency level is 65.4 percent (Census 2001). Short of what 5 percent of individuals can either read or compose English. As the vast majority of the state government works in commonplace dialects although the focal government's authority reports and reports are in English or Hindi, these records are to be deciphered into the particular common dialects to have a fitting correspondence with the individuals. Work in the region of Machine Translation in India has been continuing for a few decades. Throughout the early 90s, propelled research in the field of Artificial Intelligence and Computational Linguistics made a guaranteeing advancement of interpretation innovation. This aided in the improvement of usable Machine Translation Systems in certain decently characterized spaces. Since 1990, Scrutinize on MT frameworks between Indian and outside dialects and additionally between Indian dialects are going ahead in different organizations. Interpretation between structurally comparative dialects like Hindi and Punjabi is simpler than that between dialect matches that have wide structural distinction like Hindi and English. Interpretation frameworks between nearly related dialects are less demanding to create since they have numerous parts of their linguistic uses and vocabularies in like manner [2].

The organization of the paper is as follows. Section II gives an outline on the Assamese and Bengali language. Section III describes the related work on the machine translations. Section IV gives an outline on machine translation as well as on statistical machine translation. In section V, the design and implementation of the system has been discussed. Section VI gives the results obtained from our experiment. Section VII concludes the report.

## II. Assamese and Bengali Language

Assamese is the main language of the state Assam and is respected as the most widely used language of the entire North-East India. It is talked by most of the locals of the state of Assam. As a first language it is talked by over 15.3 million individuals and including the individuals who talk it as a second language, what added up to 20 million. Assamese is mainly used in the North-Eastern state of Assam and in parts of the neighboring states of West Bengal, Meghalaya. Little pockets of Assamese speakers can additionally be found in Bhutan and Bangladesh. Settlers from Assam have conveyed the dialect with them to different parts of the world. Although researchers follow the historical backdrop of Assamese writing to the start of the second millennium AD, yet an unbroken record of artistic history is traceable just from the fourteenth century. The Assamese dialect developed out of Sanskrit, the antiquated dialect of the Indian sub-mainland. Notwithstanding, its vocabulary, phonology and language structure have significantly been affected by the first occupants of Assam, for example, the Bodos and the Kacharis.

Bengali or Bangla is an Indo-Aryan language originated from Sanskrit. It is local to the locale of eastern South Asia known as Bengali, which embodies present day Bangladesh and the Indian state of West Bengali. With almost 230 million local speakers, Bengali is a stand out amongst the most prevalently talked languages on the planet. Bengali takes after Subject-Object-Verb word structures, in spite of the fact that varieties to this subject are basic. Bengali makes utilization of postpositions, as restricted to the prepositions utilized within English and other European dialects. Determiners take after the thing, while numerals, modifiers, and owners go before the thing. Bengali has two abstract styles: one is called Sadhubhasa (exquisite dialect) and the other Chaltibhasa (current dialect) or Cholit Bangla. The previous is the customary abstract style focused around Middle Bengali of the sixteenth century, while the latter is a twentieth century creation and is displayed on the vernacular talked in the Shantipur area in West Bengal, India.

## III. Related Works

Georgetown College in 1954 has designed the first Russian to English MT framework. After that numerous MT projects have been designed with many different qualities. Around 1970s, the centre of MT movement changed from the United States to Canada and then Europe. Then, the European Commission introduced 'Systran' which is a French-English MT framework. Around 1980s, numerous MT frameworks showed up [3]. CMU, IBM, ISI, and Google are utilization expression based frameworks with great results. In the early 1990s, the advancement made by the requisition of factual strategies to discourse distinguishment, presented by IBM scientists, was in absolutely SMT models. Today, superb programmed interpretation came into view. The entire examination group has moved towards corpus-based methods.

**Machine Translation Projects in India**

MT is a developing examination range in NLP for Indian dialects. MT has various methods for English to Indian dialects and Indian dialects to Indian dialects. Numerous scientists, and people, are included in the advancement of MT frameworks. The primary improvements in Indian dialect MT frameworks are given below:

*1. ANGLABHARTI (1991):* This is a machine aided translation system for translation between English to Hindi, for Public Health Campaigns. It analyses English only once and creates an intermediate structure that is almost disambiguated. The intermediate structure is then converted to each Indian language through a process of text generation.

*2. ANGLABHARTI -II (2004):* This system (Sinha et al., 2003) solved the disadvantages of the previous system. In order to improve the performance of translation a different approach ie, a Generalized Example-Base (GEB) for hybridization in addition to a Raw Example-Base (REB) is used. In this system a match in REB and GEB is first attempted before invoking the rule-base. Here various sub modules are pipelined which gives more accuracy. ANGLABHARTI technology is presently under the ANGLABHARTI Mission. The main aim of this is to develop Machine Aided Translation (MAT) systems for English language to twelve different Indian regional languages like Marathi and Konkani, Assamese and Manipuri, Bangla, Urdu, Sindhi and Kashmiri, Malayalam, Punjabi, Sanskrit, Oriya. [4]

*3. ANUBHARATI (1995):* This system aimed at translating Hindi to English. It is based on machine aided translation where template or hybrid HEBM is used. The HEBMT has the advantage of pattern and example-based approaches. It provides a generic model for translation between any two Indian languages pair [5].

*4. ANUBHARATI-II (2004):* ANUBHARATI-II is a reconsidered form of the ANUBHARATI that overcomes the majority of the burdens of the prior building design with a fluctuating level of hybridization of distinctive standards. The principle expectation of this framework is to create Hindi to any other Indian dialects, with a summed up various leveled case based methodology. In any case, while both ANGLABHARTI-I and ANUBHARTI-II did not prepare the normal results, both frameworks have been actualized effectively with great results. [5]

*5. MaTra (2004):* Matra is a Human-Assisted interpretation framework which converts English to Indian dialects (at present Hindi). Matra is an inventive framework where the client can examine the investigation of the framework and can give disambiguation data to prepare a solitary right interpretation. Matra is a progressing project and the framework till now can work out area particular basic sentences. Improvement has been made towards coating different sorts of sentences [5].

*6. MANTRA (1999):* This framework is mainly for English to Indian dialects and additionally from Indian dialects to English. The framework can protect the designing of data word reports over the interpretation. [5].

*7. A hybrid MT system for English to Bengali:* This MT framework for English to Bengali was created at Jadavpur University, Kolkata, in 2004. The current adaptation of the framework works at the sentence level. [6]

## IV. LITERATURE REVIEW

### A. Machine Translation

Machine Translation (MT) is the use of computers to automate the production of translations from one natural language into another, with or without human assistance. Machine translation System is used to translate the source text into target text. MT system uses the various approaches to complete the translation. Machine translation is considered as difficult task. These systems incorporate a lot of knowledge about words, and about the language (linguistic knowledge). Such knowledge is stored in one or more lexicons, and possibly other sources of linguistic knowledge, such as grammar. The lexicon is an important component of any MT system. A lexicon contains all the relevant information about words and phrases that is required for the various levels of analysis and generation. A typical lexicon entry for a word would contain the following information about the word: the part of speech, the morphological variants, the expectations of the word some kind of semantic or sense information about the word, and information about the equivalent of the word in the target language [7].

**Challenges in Machine translation:**

1) All the words in one language may not have equivalent words in another language. Sometimes a word in one language is expressed by a group of words in another.
2) Two given languages may have completely different structures. For example English has SVO structure while Assamese has SOV structure.
3) Words can have more than one meaning and sometimes group of words or whole sentence may have more than one meaning in a language.
4) Since all the natural languages are very vast so it is almost not possible to include all the words and transfer rules in a dictionary.
5) Since both Assamese and Bengali are free-word-order languages, so sometimes the translation of a sentence may give different meaning.
6) Assamese language produces negations by putting a না or ন in front of the verb. Assamese verbs have complete sets of negative conjugations with the negative particle `na-'. Bengali doesn't have any negative conjugations.
7) Assamese definitive (the Assamese for `the': টো (tu), জন (jan), জনী (jani), খন (khan) etc.) have no parallels in Bengali.

### B. Statistical Machine Translation

The Statistical Machine Translation (SMT) system is based on the view that every sentence in a language has a possible translation in another language. A sentence can be translated from one language to another in many possible ways. Statistical translation approaches take the view that every sentence in the target language is a possible translation of the input sentence [7]. Figure 1 gives the outline of the Statistical Machine Translation system.

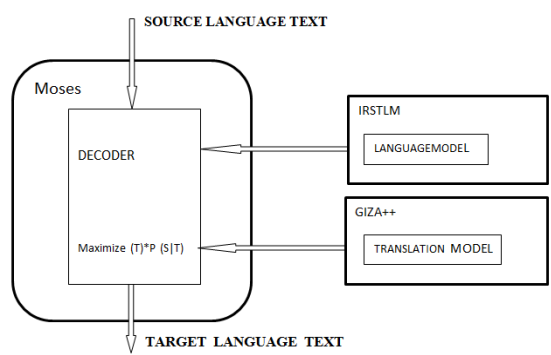

Fig 1: Architecture of SMT system

**Language Model**

A language model gives the probability of a sentence computed using n-gram model. Language model can be considered as computation of the probability of single word given all of the word that precedes it in a sentence. It is decomposed into the product of conditional probability. By using chain rule this is made possible as shown in below. The probability of sentence P(S), is broken down as the probability of individual words P(W).

P(S) = P (w1, w2, w3 ...wn)

P(W) = P(w1)P(w2|w1)P(w3|w1w2)P(w4|w1w2w3) ...P(wn|w1 w2 ...wn1)

An n-gram model simplifies the task by approximating the probability of a word given all the previous words. An n-gram of size 1 is referred to as a Unigram; size 2 is a Bigram (or, less commonly, a diagram) and so on.

**Translation Model**

This model helps to compute the conditional probability P (T|S). It is trained using the parallel corpus of target-source pairs. As no corpus is large enough to allow the computation translation model probabilities at sentence level, so the process is broken down into smaller units, e.g., words or phrases and their probabilities learnt. The translation of source sentence is thought of as being generated from source word by word. A target sentence is translated as given in figure 2.

(আসামে একটি সুন্দর জায়গা। | অসম এখন সুন্দৰ ঠাই।)

আসামে একটি সুন্দর জায়গা।
 |     |    |    |
অসম এখন সুন্দৰ ঠাই    ।

Fig 2: Translation of a sentence

Possible alignment for the pair of sentences can be represented as:

(আসামে একটি সুন্দর জায়গা । | অসম (1) এখন (2) সুন্দৰ (3) ঠাই (4) । (5))

A number of alignments are possible. For simplicity, word by word alignment of translation model is considered. The above set of alignment is denoted as *A(S, T)*. If length of target is m and that of source is n, then there are mn different alignments are possible and all connection for each target position are equally likely, therefore order words in T and S does not affect P (T|S) and likelihood of (T|S) can be defined of conditional probability P (T, a/S) shown as P (S|T) = sum P(S, a/T). The sum is over the element of alignment set, A(S, T).

**Decoder**

This phase of SMT maximizes the probability of translated text. The words are chosen which have maximum like hood of being the translated translation. Search for sentence T is performed that maximizes P(S|T) i.e. Pr(S, T) = argmax P(T) P(S|T). Here problem is the infinite space to be searched. The use of stacked search is suggested, in which we maintain a list of partial alignment hypothesis. Search starts with null hypothesis, which means that the target sentence is obtained from a sequence of source words that we do not know. [7]

## V. METHODOLOGY

This section includes corpus collection, data preparation, development of Language Model, Translation Model and training of decoder using Moses tool.

### A. Corpus Preparation

Statistical Machine Translation system uses a parallel corpus of source and target language pairs. For this, we have developed a Bengali to Assamese parallel corpus with approx. 20000 sentences. This corpus consists of small sentences related to novel, story, travel, tourism in India. Table \ref{corpus} shows the number of sentences use in training, testing and tuning purposes.

| Corpus | No of sentences | Source | Target |
|---|---|---|---|
| Training | 17000 | 17000 | 17000 |
| Testing | 1500 | 1500 | 1500 |
| Tuning | 1500 | 1500 | 1500 |

TABLE 1: No of sentences for training, testing and tuning.

### B. Language Model Training

Language model are created by a language modeling toolkit. The model is created with a number of variables that can be adjusted to enable better translation. The models are building with the target language (i.e. Assamese) as it is important for it to know the language it outputs should be structured. The IRSTLM documentation gives a full explanation of the command-line option [8].

### C. Training the Translation System

Finally we come to our important phase - training the translation model. This will run word alignment (using GIZA++), phrase extraction and scoring, create lexicalized reordering tables and create our Moses configuration file. We have created an appropriate directory as follows, and then run the training command logs [8]:

mkdir /work

cd /work

nohup

nice /mymoses/scripts/training/train-model. perl-root-dirtraincorpus

/corpusproject/ben-ass1.clean

-f as-e en –alignment grow-diag-final-and

–reordering msd-bidirectional-fe

-lm 0:3: $HOME/lm/ben-ass1.blm.ben:8

-external-bin-dir /mymoses/tools > &training.out

Once it is finished, a *moses.ini* file will be created in the directory */work/train/model*. We can use this ini file to decode, but there are a couple of problems with it. The first is that it is very slow to load (usually in case of large corpus), but we can fix that by bin-arising the phrase table and reordering table, i.e. compiling them into a format that can be load quickly. The second problem is that the weight used by Moses to weight the different models against each other are not optimized –if we look at the moses.ini file we see that they are set to default values like 0.2, 0.3, etc. To find better weights we need to tune the translation system, which leads us to the next step.

### D. Tuning

Tuning is the slowest part of the process. We have again collected a small amount of parallel data separate from the training data [8]. We are going to tokenize and truecase it first, just as we did the training process. Now we again go back to the training directory and launched the tuning process. After the tuning process is finished, an *ini* file is created with train weights, which is in the directory ~/work/mert-work/moses.ini.

### E. Testing

We can now run moses with the following command:

~/my moses/bin/moses –f \~/work/mart-work/moses.ini

We can type now one Bengali sentence and get the output in Assamese. We can also echo the sentence to get the output like this:

Echo "আমি গৌহাটি বিশ্ববিদ্যালয়ের একজন ছাত্র" | ~/my moses/bin/moses -f ~/work/mert-work/moses.ini

This will give the output: "মই গুৱাহাটী বিশ্ববিদ্যালয়ৰ ছাত্ৰ"

We can now measure how good our translation system is. For this, we use another parallel data set. So we again tokenize and truecase it as before. The model that we have trained can then be filtered for this test set; meaning that we only retain the entries needed translate the test set. This will make the translation a lot faster. We can test the decoder by translating the test set then running the BLEU scripts on it.

## VI. RESULT AND ANALYSIS

Table II shows some translating sentences in our system.

| Bengali sentence translating | Assamese sentence |
|---|---|
| দিল্লী ভারতের রাজধানী | দিল্লী ভাৰতৰ ৰাজধানী |
| আসামে একটি সুন্দর জায়গা | অসম এখন সুন্দৰ ঠাই |
| ভারত একটি বড় দেশ | ভাৰত এখন ডাঙৰ দেশ |
| হায়দরাবাদ অন্ধ্রপ্রদেশ মধ্যে অবস্থিত | হায়দৰাবাদ অন্ধ্রপ্রদেশৰ মধ্যত অৱস্থিত |
| উদয়পুর রাজস্থানে দক্ষিণ অংশে অবস্থিত | উদয়পুৰ ৰাজস্থানৰ দক্ষিণ অংশত অৱস্থিত |
| আমি গৌহাটি বিশ্ববিদ্যালয়ের একজন ছাত্র | মই গুৱাহাটী বিশ্ববিদ্যালয়ৰ ছাত্ৰ |

TABLE II: Some translating sentences in our system

For the experiments, we have chosen three sets of randomly selected sentences with 200, 250 and 300 sentences. Table III is the analysis table for the observation of set of sentences. We have graphically shown the values in Figure 3. In the wake of experiencing the results, we can say that the slips are a direct result of the emulating reasons:

1. The amount of words in our corpus is extremely constrained.
2. The PoS tagger sections are not finished.
3. Now and then, due to different word passages in the target dialect lexicon for a solitary word in the source dialect lexicon. For instance, for both the Assamese words নগৰ and চহৰ, the Bengali word is শহর

| Sets | Total | Successful | Unsuccessful | % of error |
|---|---|---|---|---|
| Set 1 | 200 | 165 | 35 | 17.5 |
| Set 2 | 250 | 211 | 39 | 15.6 |
| Set 3 | 300 | 259 | 41 | 13.7 |

TABLE III: Analysis table for observation of sentences

The output of the experiment was evaluated using BLEU (Bilingual Evaluation Understudy). BLEU toolkit is used to calculate the BLEU (Determine how good the translation is, the metric estimates the quality of machine translated text) [8]. We obtained a BLEU score of 16.3 from the parallel corpus (Bengali-Assamese) after translation. This is very small and may be because we have used a very small data set. BLEU score are not commensurate even between different corpora in the same translation direction. BLEU is really only comparable for different system or system variant on the exact same data. In the case of same corpus in two directions, an imperfect analogy might be gas mileage between two different cities. No matter how consistently you drove, you would not expect the same gasoline usage driving from city C to D as in the other direction, (especially if one direction is more uphill than the other).

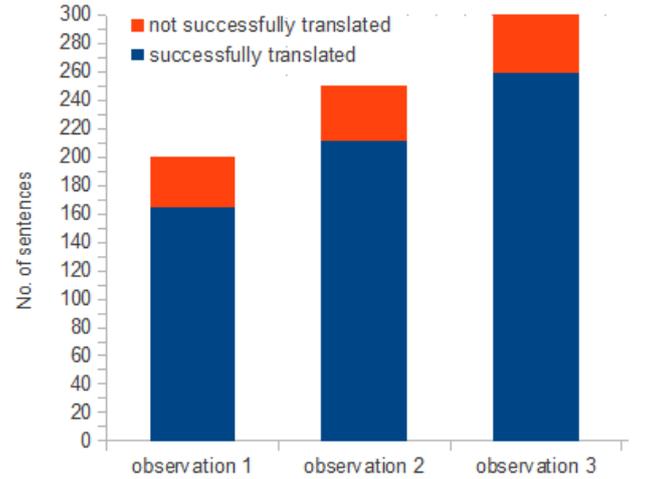

Fig 3: Graphical analysis for observation of sentences

## VII. CONCLUSIONS AND FUTURE WORK

In this paper, Bengali to Assamese SMT system has been developed. Our method of extracting translation pattern is a relatively simple one and has been proposed by using phrase-based decoder in Moses. We extracted the results of BLEU score. We will try to develop the translation system by our own instead of using Moses MT system. We will try to increase the corpus for better training for better efficiency. The system can also be put in the web-based portal to translate content of one web page in Bengali to Assamese. We try to get more corpuses from different domains in such a way that it will cover all the wording. Since BLEU is not good we need some evaluation techniques also. We should try the incorporation of shallow syntactic information (POS tags) in our discriminative model to boost the performance of translation.